\documentclass[runningheads]{llncs}

\usepackage{accv}

\usepackage{accvabbrv}

\usepackage{graphicx}
\usepackage{booktabs}
\usepackage{rotating}
\usepackage{algorithmic}
\usepackage{algorithm}

\usepackage[accsupp]{axessibility}  

\usepackage{hyperref}

\usepackage{orcidlink}

\begin{document}

\title{Feature Space Analysis by Guided Diffusion Model} 


\author{Kimiaki Shirahama\orcidlink{0000-0003-1843-5152} \and Kaduki Yamashita \and Miki Yanobu \and Miho Ohsaki\orcidlink{0000-0002-6766-2928}}

\authorrunning{K.~Shirahama et al.}

\institute{
	Doshisha University, 1-3 Tatara Miyakodani, Kyotanabe, Kyoto, 610-0394, Japan\\
	\email{\{kshiraha,mohsaki\}@mail.doshisha.ac.jp}
}

\maketitle

\begin{abstract}
This paper aims to analyse the feature space of a vision-related Deep Neural Network (DNN) by proposing a \textit{decoder} that can generate an image whose feature closely matches a user-specified feature. Supported by quantitative evidence of its high feature-matching accuracy, our decoder facilitates precise analysis of the DNN's feature space. Our decoder is implemented as a \textit{guided diffusion model} that guides the image generation of a pre-trained diffusion model to minimise the Euclidean distance between the feature of a clean image estimated at each step and the user-specified feature. The key advantages of our decoder are its training-free applicability to analyse the feature spaces of different DNNs and its practical feasibility on a single COTS GPU. The experiments targeting CLIP's image encoder and ResNet-50 demonstrate the effectiveness of our decoder both as a feature-matching image generator and as a visual feature space analyser. The codes and data are available at \url{https://github.com/ccilab-doshisha/FeatDec}.
\end{abstract}

\section{Introduction}
\label{sec:intro}

Deep learning is often referred to as representation learning or feature learning that extracts useful features through statistical analysis on a large amount of data~\cite{Y_Bengio,F_Li}. In vision-related domains, features extracted by a Deep Neural Network (DNN) not only yield accurate image classification~\cite{K_He2,A_Dosovitskiy} but also serve as building blocks for various tasks such as cross-modal retrieval~\cite{F_Faghri,A_Radford},  image/video captioning~\cite{J_Donahue} and image editing~\cite{P_Isola,A_Hertz}. However, despite these extensive explorations of utilising features, very limited attention has been given to analyse the underlying feature space. Most studies only examine distributional characteristics of features by visualising them with dimensionality reduction methods like t-SNE and UMAP~\cite{M_Laurens,L_McInnes,S_Damrich}. Another motivation arises from visual-semantic embedding, in which an image and its relevant textual caption are encoded into similar features in a common space~\cite{F_Faghri,A_Radford}, but they are not the same. This raises the question of what kind of image (or caption) is characterised by a feature located between those of the relevant image and caption. 

To analyse the feature space of a DNN, we develop a \textit{decoder} that generates an image whose feature closely matches a user-specified feature. Since the user can specify arbitrary features in the DNN's feature space, our decoder allows us to visually interpret a wide range of feature-image correspondences. Our decoder is implemented as a \textit{guided diffusion model} where a pre-trained diffusion model serves as a generic high-quality image generator~\cite{J_Ho,R_Rombach} and its image generation is guided to minimise a loss function~\cite{A_Bansal,A_Bansal2}. Our decoder's loss function measures the Euclidean distance between a user-specified feature and the feature of a generated image, thereby encouraging the latter to closely match the former. With precise matching quantified by the small Euclidean distances between user-specified features and those of generated images, our decoder facilitates precise analysis of the DNN's feature space. To the best of our knowledge, prior work has not addressed this kind of precise feature space analysis.



The experiments to analyse the feature spaces of CLIP's image encoder~\cite{A_Radford} and ResNet-50~\cite{K_He2} demonstrate the effectiveness of our decoder from the perspectives of both feature-matching image generation and visual feature space analysis. Finally, our decoder has the potential to open up new directions in image mining. For example, a ``rare'' image can be generated by specifying a feature that lies far from the feature of every image in a dataset. A ``memorised'' image, for which a small perturbation of its feature leads to a different prediction~\cite{R_Bansal}, can be generated by specifying a feature on or near the decision boundary of a DNN~\cite{S_Yang}. A ``hard negative'' image, whose feature is difficult to distinguish from the one extracted from a semantically different image~\cite{F_Faghri}, can be identified by repeatedly generating images with varied user-specified features and applying image classification to those images. These images are expected to be useful for improving the feature space of a DNN. To mine such images, one of the primary focuses of this paper is to pursue a high feature-matching accuracy that enables fine-grained control of image generation via user-specified features.

\section{Related Work}

This section reviews existing studies from the perspectives of guided diffusion models and decoders. Many guidance approaches have been developed to provide different controls over the image generation of a diffusion model in order to devise various applications such as text-to-image~\cite{P_Dhariwal,J_Ho2,A_Nichol,H_Chefer,G_Kim,K_Sueyoshi,A_Bansal,A_Bansal2}, image inpainting~\cite{A_Nichol,A_Bansal,A_Bansal2}, object/style/layout manipulation~\cite{A_Graikos,D_Epstein,A_Bansal,A_Bansal2}, composition between images~\cite{D_Epstein}, and image segmentation~\cite{A_Graikos}. The most basic loss functions like classifier guidance~\cite{P_Dhariwal}, classifier-free guidance~\cite{J_Ho2} and CLIP guidance~\cite{A_Nichol} measure a probability or score indicating how relevant a text prompt is to a noisy image generated at each step. These loss functions are then extended to examine whether each of subject tokens in a prompt is attended to by some patches in a generated image~\cite{H_Chefer}, whether a generated image has a plausible depth map~\cite{G_Kim} and whether a generated image satisfies propositional statements deduced from a prompt~\cite{K_Sueyoshi}. In addition, various loss functions are proposed to assess different aspects such as an object's position, size, shape and appearance~\cite{D_Epstein}, an object's pose and position, personal identity and painting style~\cite{A_Bansal,A_Bansal2} and the color and label distributions in a generated segmentation image~\cite{A_Graikos}.


To the best of our knowledge, existing studies on guided diffusion models have not specifically focused on generating images whose features closely match a user-specified feature. Our approach is inspired by \cite{A_Bansal,A_Bansal2} that demonstrate the effectiveness of applying a loss function to a clean image, which is analytically predicted from the noisy image generated at each step of the reverse process. In other words, such clean images are naturally imperfect but still useful for providing appropriate guidance to minimise the loss function. By borrowing this idea, we propose a guided diffusion model that employs a loss function designed to compute the Euclidean distance between a user-specified feature and the one extracted from a clean image predicted at each step.

Decoders that invert a feature back into an image from which the same feature is likely to be extracted have been studied in feature inversion. Existing decoders are categorised as training-based or training-free, depending on whether they require training. Training-based decoders are typically designed as conditional generative models that are conditioned on a user-specified feature to generate an image, which is likely to be encoded into this feature~\cite{P_Teterwak,A_Ramesh,F_Bordes}. However, training-based decoders involve two crucial problems: First, just using a user-specified feature as conditioning lacks checking how close the feature of a generated image is to the user-specified one. The experiment in Section~\ref{ssec:result_resnet} shows that although images generated by the training-based decoder in \cite{F_Bordes} are plausible, their features are distant from the user-specified features. Second, training generative models on large-scale datasets requires very expensive computational costs. In particular, a diffusion model should be employed as a decoder because of the quality of generated images, but its training is prohibitively expensive for most researchers. For instance, training the diffusion model in \cite{P_Dhariwal} consumes $150$-$1000$ V100 GPU days to produce high-quality images~\cite{Z_Wang}. Moreover, training stable diffusion~\cite{R_Rombach} demands over $24$ days on $256$ A100 GPUs~\cite{H_Zheng}. 

Training-free decoders like ours circumvent computationally expensive training. Classical decoders based on direct pixel optimisation towards a user-specified feature~\cite{A_Mahendran,C_Olah} clearly struggle to generate realistic images due to the lack of a strong image prior. Our decoder uses a pre-trained diffusion model as a powerful image prior. Also, training-free decoders have been implemented in some recent studies~\cite{S_Fu,S_Zhang,A_DOrazio} to demonstrate objectives that differ from ours, namely human-aligned perceptual similarity metric~\cite{S_Fu}, secure transmission of early-layer features~\cite{S_Zhang} and new image generation framework~\cite{A_DOrazio}. Consequently, these studies do not quantitatively evaluate how accurately the features of generated images match user-specified ones. In contrast, we aim at precise feature space analysis and explicitly evaluate the feature-matching accuracy of our decoder. Finally, compared to resource-intensive training-based decoders, our training-free decoder only requires a single COTS GPU with more than $16$ GB of VRAM.



\section{Our Guided-diffusion-model-based Decoder}
\label{sec:our_method}



\begin{figure}[tb]
	\centering
	\includegraphics[width=0.75\linewidth]{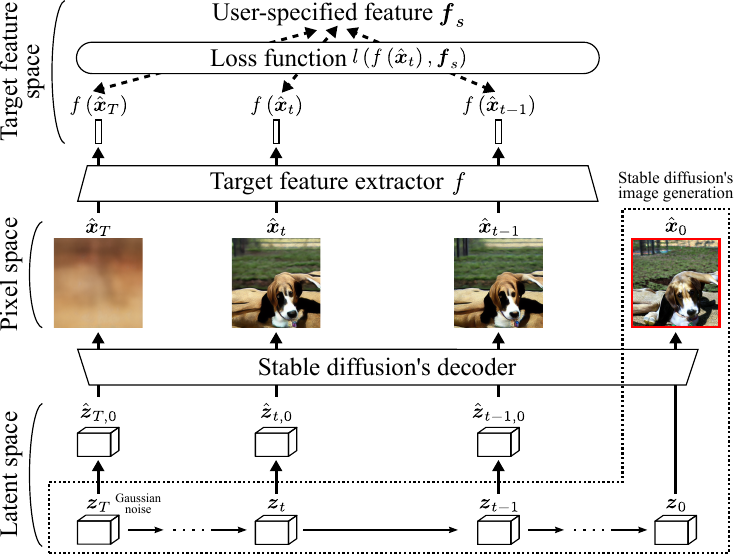}
	\caption{An overview of our decoder that guides stable diffusion's image generation to generate an image whose feature closely matches a user-specified feature $\boldsymbol{f}_s$. At step $t$, our decoder estimates a clean latent representation $\hat{\boldsymbol{z}}_{t,0}$ from $\boldsymbol{z}_{t}$, generates the corresponding image $\hat{\boldsymbol{x}}_t$ from $\hat{\boldsymbol{z}}_{t,0}$, extracts the feature $f \left( \hat{\boldsymbol{x}}_t \right)$ of $\hat{\boldsymbol{x}}_t$, and computes the loss as the squared Euclidean distance between $f \left( \hat{\boldsymbol{x}}_t \right)$ and $\boldsymbol{f}_s$.}
	\label{fig:overview}
	\vspace{-0.5cm}
\end{figure}

\noindent \textbf{Overview of our decoder:} Fig.~\ref{fig:overview} shows an overview of our decoder consisting of the following three main components: The first is a pre-trained diffusion model, especially stable diffusion~\cite{R_Rombach}\footnote{The exact checkpoint is sd-v1-4.ckpt available on \url{https://huggingface.co/CompVis/stable-diffusion-v-1-4-original}.} illustrated by the dotted region, that establishes the foundational image generation process. The second component depicted by the trapezium is the target feature extractor $f$ whose feature space is our analysis target. A guidance signal for controlling stable diffusion's image generation is computed in $f$'s feature space via the loss function $l \left( f \left( \hat{\boldsymbol{x}}_t \right), \boldsymbol{f}_s \right)$, which is the last component expressed by the rectangle with rounded corners. Taking into account these three main components, our decoder is detailed below.

Our decoder is an extension of the guided diffusion model proposed in \cite{A_Bansal,A_Bansal2}\footnote{Based on our preliminary experiments, the backward universal guidance module, one of the main modules in \cite{A_Bansal,A_Bansal2}, is not adopted in our decoder. This module optimises image pixels without considering the step progress of the reverse process, often yielding unrealistic or corrupted images with limited and inconsistent loss reduction.}. In stable diffusion, the reverse process is performed in the latent space built by a pre-trained autoencoder in order to attain perceptually equivalent but computationally efficient image generation. Specifically, a latent representation $\boldsymbol{z}_T$ (to be precise, a feature map of shape $4 \times 64 \times 64$) defined by pure Gaussian noise is progressively denoised into its clean version $\boldsymbol{z}_0$, from which the image $\hat{\boldsymbol{x}}_0$ is generated through the autoencoder's decoder in stable diffusion (hereafter, stable diffusion's decoder).

Our guidance modulates each reverse step to sample $\boldsymbol{z}_{t-1}$ from $\boldsymbol{z}_t$ ($1 \leq t \leq T$). The core idea is that a clean latent representation $\hat{\boldsymbol{z}}_{t,0}$ which is imperfect but still useful can be estimated as follows~\cite{A_Bansal,A_Bansal2}:
\begin{align}
	\hat{\boldsymbol{z}}_{t,0} = \frac{\boldsymbol{z}_t - \sqrt{1 - \bar{\alpha}_t} \ \epsilon_{\theta} \left( \boldsymbol{z}_t, t \right)}{ \sqrt{\bar{\alpha_t}} }.
	\label{eq:z0_pred}
\end{align}
This equation is derived based on the forward process to create $\boldsymbol{z}_t = \sqrt{\bar{\alpha_t}} \boldsymbol{z}_0 + \sqrt{1 - \bar{\alpha}_t} \ \boldsymbol{\epsilon}$ where $\boldsymbol{\epsilon} \sim \mathcal{N} \left(\boldsymbol{0}, \mbox{\textbf{I}}\right)$ is source noise and $\bar{\alpha}_t = \prod_{i=1}^{t} \alpha_i = \prod_{i=1}^{t} (1 - \beta_i)$ and $1 - \bar{\alpha}_t$ respectively indicate the scale of clean $\boldsymbol{z}_0$ and that of $\boldsymbol{\epsilon}$ according to the pre-defined variance schedule $\beta_1, \cdots, \beta_t$. In Eq. (\ref{eq:z0_pred}), $\boldsymbol{z}_0$ is approximated as $\hat{\boldsymbol{z}}_{t,0}$ using $\epsilon_{\theta} \left( \boldsymbol{z}_t, t \right)$ that is source noise predicted by passing $\boldsymbol{z}_t$ and $t$ to stable diffusion's UNet parameterised by $\theta$~\footnote{Following \cite{A_Bansal,A_Bansal2}, an empty string is also passed to implement unconditional image generation using text-conditional stable diffusion.}.

As illustrated in Fig.~\ref{fig:overview}, the image $\hat{\boldsymbol{x}}_t$ is then generated by feeding $\hat{\boldsymbol{z}}_{t,0}$ into stable diffusion's decoder. Thereby, according to the target feature extractor $f$, it is possible to examine how close the feature $f \left( \hat{\boldsymbol{x}}_t \right)$ of $\hat{\boldsymbol{x}}_t$ is to a user-specified feature $\boldsymbol{f}_s$ in terms of their squared Euclidean distance below:
\begin{align}
	l \left( f \left( \hat{\boldsymbol{x}}_t \right), \boldsymbol{f}_s \right) = \left|\left| f \left( \hat{\boldsymbol{x}}_t \right) - \boldsymbol{f}_s \right|\right|_2^2.
	\label{eq:euc_loss}
\end{align}
Our guidance uses $l \left( f \left( \hat{\boldsymbol{x}}_t \right), \boldsymbol{f}_s \right)$ as a loss function to sample $\boldsymbol{z}_{t-1}$ that is likely to yield $\hat{\boldsymbol{z}}_{t-1,0}$, followed by the image $\hat{\boldsymbol{x}}_{t-1}$ whose feature $f \left( \hat{\boldsymbol{x}}_{t-1} \right)$ is closer to $\boldsymbol{f}_s$ than $f \left( \hat{\boldsymbol{x}}_t \right)$. Specifically, the predicted source noise $\epsilon_{\theta} \left( \boldsymbol{z}_t, t \right)$ for $\boldsymbol{z}_t$ is modified by $\nabla_{\boldsymbol{z}_t} l \left( f \left( \hat{\boldsymbol{x}}_t \right), \boldsymbol{f}_s \right)$ that is the gradient of $l \left( f \left( \hat{\boldsymbol{x}}_t \right), \boldsymbol{f}_s \right)$ in terms of $\boldsymbol{z}_t$:
\begin{align}
	\epsilon_{\theta}' \left( \boldsymbol{z}_t, t \right) = \epsilon_{\theta} \left( \boldsymbol{z}_t, t \right) + w_g  \nabla_{\boldsymbol{z}_t} l \left( f \left( \hat{\boldsymbol{x}}_t \right), \boldsymbol{f}_s \right),
	\label{eq:esp_modify}
\end{align}
where $w_g$ is a hyper-parameter to balance $\epsilon_{\theta} \left( \boldsymbol{z}_t, t \right)$ and $\nabla_{\boldsymbol{z}_t} l \left( f \left( \hat{\boldsymbol{x}}_t \right), \boldsymbol{f}_s \right)$.

Finally, $\boldsymbol{z}_{t-1}$ is sampled by the following DDPM sampler~\cite{J_Ho} except that $\epsilon_{\theta} \left( \boldsymbol{z}_t, t \right)$ is replaced with $\epsilon_{\theta}' \left( \boldsymbol{z}_t, t \right)$:
\begin{align}
	\boldsymbol{z}_{t-1} = \frac{1}{\sqrt{\alpha_t}} \left( \boldsymbol{z}_t - \frac{1 - \alpha_t}{\sqrt{1 - \bar{\alpha}}_t} \epsilon_{\theta}' \left( \boldsymbol{z}_t, t \right) \right) + \sqrt{\beta_t} \boldsymbol{\Delta},
	\label{eq:z_sampling}
\end{align}
where $\boldsymbol{\Delta}$ is noise sampled from $\mathcal{N} \left(\boldsymbol{0}, \mbox{\textbf{I}}\right)$. Compared to fast samplers like DDIM~\cite{J_Song}, DDPM is employed to allow for many steps so as to minimise the loss function as much as possible~\cite{D_Watson,V_Tong}. By repeating the above-mentioned sampling of $\boldsymbol{z}_{t-1}$ from $\boldsymbol{z}_{t}$, our guided diffusion model generates an image $\hat{\boldsymbol{x}}_0$ whose feature $f \left( \hat{\boldsymbol{x}}_0 \right)$ minimises the squared Euclidean distance to $\boldsymbol{f}_s$.


\begin{algorithm}[htb]
	\caption{Our decoder's guided image generation}
	\label{alg:decoder}
	\begin{algorithmic}[1]
		\REQUIRE Target feature extractor $f$, stable diffusion's UNet $\epsilon_{\theta}$ and decoder, number of reverse steps $T$, variance schedule $\left\{ \beta_1, \cdots, \beta_T \right\}$, user-specified feature $\boldsymbol{f}_s$, gradient weight $w_g$
		\STATE Sample $\boldsymbol{z}_{T}$ from $\mathcal{N} \left(\boldsymbol{0}, \mbox{\textbf{I}}\right)$
		\FOR{$t= T, \cdots 1$}
		\STATE Set the number of self-recurrence iterations $K$ based on the early step emphasis strategy 
		\FOR{$k=1, \cdots, K$}
		\STATE Compute $\hat{\boldsymbol{z}}_{t,0}$ from $\boldsymbol{z}_{t}$ using Eq. (\ref{eq:z0_pred})
		\STATE Generate $\hat{\boldsymbol{x}}_t$ from $\hat{\boldsymbol{z}}_{t,0}$ using stable diffusion's decoder 
		\STATE Compute the feature $f \left( \hat{\boldsymbol{x}}_t \right)$ of $\hat{\boldsymbol{x}}_t$
		\STATE Compute $l \left( f \left( \hat{\boldsymbol{x}}_t \right), \boldsymbol{f}_s \right)$ using Eq. (\ref{eq:euc_loss}) and its gradient $\nabla_{\boldsymbol{z}_t} l \left( f \left( \hat{\boldsymbol{x}}_t \right), \boldsymbol{f}_s \right)$
		\STATE Normalise and clip  $\nabla_{\boldsymbol{z}_t} l \left( f \left( \hat{\boldsymbol{x}}_t \right), \boldsymbol{f}_s \right)$
		\STATE Compute $\epsilon_{\theta}' \left( \boldsymbol{z}_t, t \right)$ using Eq. (\ref{eq:esp_modify})
		\STATE Sample $\boldsymbol{z}_{t-1}$ using Eq. (\ref{eq:z_sampling})
		\IF{$k < K$} \STATE{Resample $\boldsymbol{z}_t$ from $\boldsymbol{z}_{t-1}$ using Eq. (\ref{eq:self_rec})} \ENDIF
		\ENDFOR
		\ENDFOR
		\STATE Generate $\hat{\boldsymbol{x}}_0$ from $\boldsymbol{z}_0$ using stable diffusion's decoder
		\RETURN $\hat{\boldsymbol{x}}_0$
	\end{algorithmic}
\end{algorithm}

\noindent \textbf{Details of our decoder:} Algorithm~\ref{alg:decoder} shows an operation flow in our decoder. The main operations at lines $1$, $5$-$8$, $10$, $11$ and $17$ have been already described before. The following part details the other operations such as early step emphasis at line $3$, self-recurrence at lines $4$, $12$ and $13$, and gradient normalisation and clipping at line $9$.

Self-recurrence is motivated by the fact that even the large number of steps $T$ in DDPM sampler (especially $T = 1000$) are insufficient to generate an image whose feature is within acceptable proximity to $\boldsymbol{f}_s$. To overcome this, self-recurrence~\cite{A_Lugmayr,Y_Wang,A_Bansal,A_Bansal2} is employed to resample $\boldsymbol{z}_t$ from $\boldsymbol{z}_{t-1}$ by following the standard forward step below:
\begin{align}
	\boldsymbol{z}_{t} = \sqrt{\alpha_t} \boldsymbol{z}_{t-1} + \sqrt{1 - \alpha_t} \boldsymbol{\Delta} \; \; \; \; \mbox{where } \boldsymbol{\Delta} \sim \mathcal{N} \left(\boldsymbol{0}, \mbox{\textbf{I}}\right).
	\label{eq:self_rec}
\end{align}
Since this resampled $\boldsymbol{z}_t$ preserves the information in $\boldsymbol{z}_{t-1}$, this $\boldsymbol{z}_t$ is likely to yield an image whose feature is closer to $\boldsymbol{f}_s$ than $\boldsymbol{z}_t$ before self-recurrence. This way, $\boldsymbol{z}_t$ is iteratively refined by incorporating the information from the subsequent $(t-1)$th step while maintaining the noise scale at the current $t$th step.

The overall object layout of a generated image is primarily established during early reverse steps~\cite{J_Ho}. If this layout is inadequate, our decoder's image generation fails. To address this, we design an early step emphasis strategy to prioritise early reverse steps. For the first $T'$ reverse steps, the number of self-recurrence iterations $K$ is set to a large value ($K=1000$ for the first $T'=5$ steps) while using a much smaller value ($K=8$) at the remaining steps.



One crucial issue to stabilise our decoder's image generation is how to balance $\epsilon_{\theta} \left( \boldsymbol{z}_t, t \right)$ and $\nabla_{\boldsymbol{z}_t} l \left( f \left( \hat{\boldsymbol{x}}_t \right), \boldsymbol{f}_s \right)$ to compute $\epsilon_{\theta}' \left( \boldsymbol{z}_t, t \right)$ in Eq. (\ref{eq:esp_modify}). Another key issue is that large values in $\nabla_{\boldsymbol{z}_t} l \left( f \left( \hat{\boldsymbol{x}}_t \right), \boldsymbol{f}_s \right)$ excessively distort the originally predicted values in $\epsilon_{\theta} \left( \boldsymbol{z}_t, t \right)$, leading to corrupted images. For the first issue, $\nabla_{\boldsymbol{z}_t} l \left( f \left( \hat{\boldsymbol{x}}_t \right), \boldsymbol{f}_s \right)$ is normalised by multiplying $\frac{|| \epsilon_{\theta} \left( \boldsymbol{z}_t, t \right) ||_2}{||\nabla_{\boldsymbol{z}_t} l \left( f \left( \hat{\boldsymbol{x}}_t \right), \boldsymbol{f}_s \right)||_2}$ where $||\boldsymbol{a}||_2$ is the L2 norm computed by flatting the tensor $\boldsymbol{a}$ into a vector. The normalised $\nabla_{\boldsymbol{z}_t} l \left( f \left( \hat{\boldsymbol{x}}_t \right), \boldsymbol{f}_s \right)$ has the same contribution as $\epsilon_{\theta} \left( \boldsymbol{z}_t, t \right)$ in terms of their L2 norms. For the second issue, we found that values in $\nabla_{\boldsymbol{z}_t} l \left( f \left( \hat{\boldsymbol{x}}_t \right), \boldsymbol{f}_s \right)$ approximately follow a Gaussian distribution whose mean is nearly zero, as demonstrated in Part A of the supplementary materials. Based on this, gradient clipping is designed such that if values in $\nabla_{\boldsymbol{z}_t} l \left( f \left( \hat{\boldsymbol{x}}_t \right), \boldsymbol{f}_s \right)$ fall outside the range $\mu \pm \nabla_{thres}$, they are clipped to their nearest boundary, either $\mu + \nabla_{thres}$ or $\mu - \nabla_{thres}$. Here, $\mu$ is the mean of values in $\nabla_{\boldsymbol{z}_t} l \left( f \left( \hat{\boldsymbol{x}}_t \right), \boldsymbol{f}_s \right)$, and $\nabla_{thres}$ is set to three times their standard deviation. Finally, $\nabla_{\boldsymbol{z}_t} l \left( f \left( \hat{\boldsymbol{x}}_t \right), \boldsymbol{f}_s \right)$ fixed by the above-mentioned gradient normalisation and clipping is weighted by $w_g=4$ to compute $\epsilon_{\theta}' \left( \boldsymbol{z}_t, t \right)$ in Eq. (\ref{eq:esp_modify}).




\section{Experimental Results}
\label{sec:exp}

We evaluate our decoder from two perspectives: The first is the evaluation as a feature-matching image generator, specifically whether our decoder can successfully generate an image whose feature closely matches a user-specified feature $\boldsymbol{f}_s$. The second perspective is the evaluation as a visual feature space analyser, namely whether a generated image provides meaningful visual cues encoded in $\boldsymbol{f}_s$. Thus, we define $\boldsymbol{f}_s$ as the feature that is extracted from an actual image via the target feature extractor $f$, because we can interpret the adequacy of visual content in a generated image by referring to the actual image.

The experiments below target two feature extractors: The first is CLIP's image encoder with ResNet-50 backbone~\cite{A_Radford} because of its popularity and stability. The second is ResNet-50 to perform a comparative study against an existing training-based decoder~\cite{F_Bordes}. Additional results for Vision Transformer (ViT)~\cite{A_Dosovitskiy} are provided in Part B of the supplementary materials.

By setting $f$ to either CLIP's image encoder or ResNet-50, $\boldsymbol{f}_s$ is extracted from each of $200$ actual images. The first and second halves of these images are randomly sampled from MS-COCO dataset (118K train and 5K 2017 validation images in the $2017$ version)~\cite{T_Lin} and ImageNet (1.2M train and 50K validation images in imagenet-1k)~\cite{O_Russakovsky} dataset, respectively. All the experiments are conducted by running our decoder on a single GPU (NVIDIA GeForce RTX 3090 ($24$GB VRAM), RTX 2000 Ada ($16$GB VRAM), or RTX 4000 Ada ($20$GB VRAM)). The hyper-parameters specific to our decoder ($K$, $T'$ and $w_g$) are fixed across all the experiments, as defined in Section~\ref{sec:our_method}. The configuration of stable diffusion is the same as the one in \cite{A_Bansal,A_Bansal2}. Finally, when clear from the context, we sometimes treat an image as its corresponding feature to simplify the expression of the Euclidean distance. For example, ``the Euclidean distance between image features'' and  ``the Euclidean distance between the feature of an image and $\boldsymbol{f}_s$'' are simplified to ``the Euclidean distance between images'' and ``the Euclidean distance between an image and $\boldsymbol{f}_s$'', respectively.




\subsection{Results Targeting CLIP's Image Encoder}
\label{ssec:result_clip}

\begin{figure}[tbp]
	\centering
	\includegraphics[width=\linewidth]{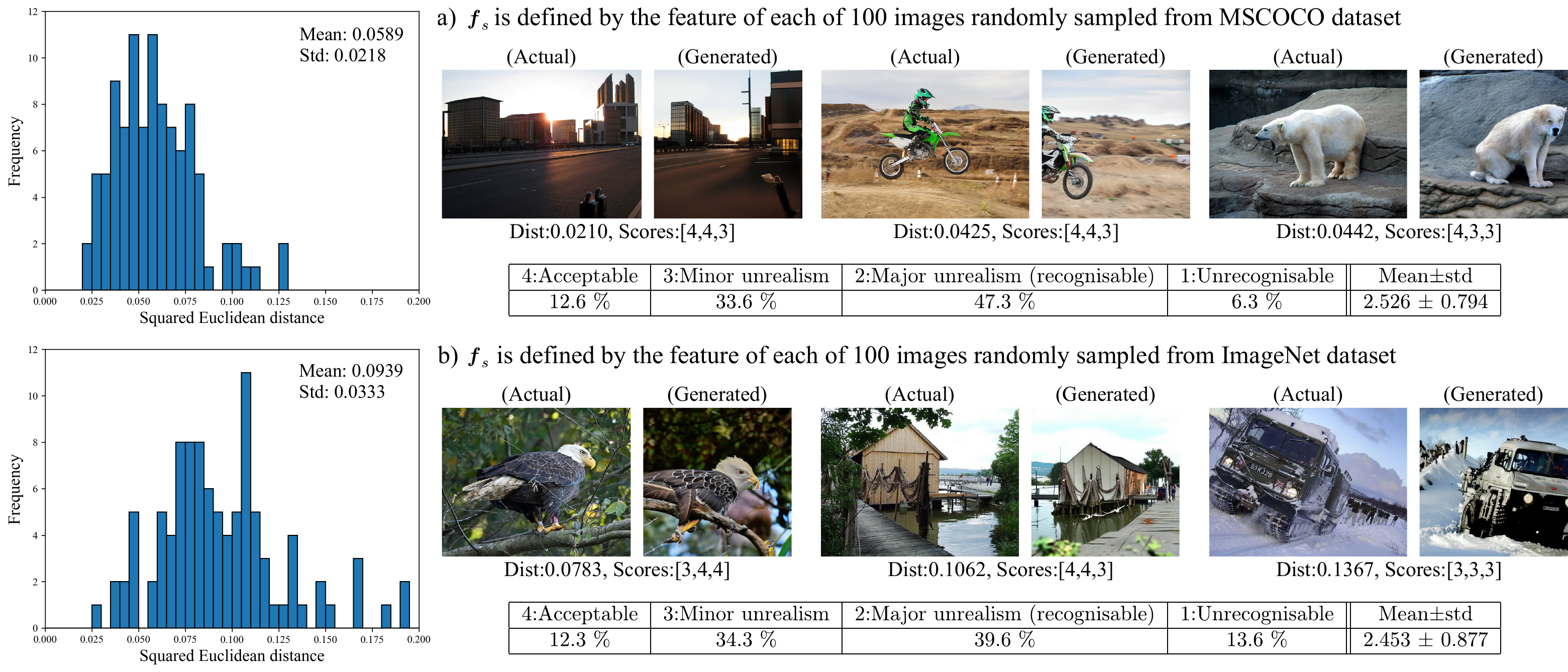}
	\caption{Results obtained by targeting CLIP's image encoder with ResNet-50 backbone and defining $\boldsymbol{f}_s$ as a feature, which is extracted from each of $100$ MS-COCO images (a) and $100$ ImageNet images (b).}
	\label{fig:gen_result_clip}
\end{figure}

Targeting CLIP's image encoder with ResNet-50 backbone, Fig.~\ref{fig:gen_result_clip} (a) and (b) summarise the results when $\boldsymbol{f}_s$ is defined as the feature extracted from each of $100$ MS-COCO images and $100$ ImageNet images, respectively. Fig.~\ref{fig:gen_result_clip} (a) and (b) both consist of three parts: The first is a histogram representing the distribution of the squared Euclidean distances between the $100$ generated images and the corresponding $\boldsymbol{f}_s$. Note that since our goal is to generate images whose features closely match $\boldsymbol{f}_s$, each run is represented by the ``best'' image that yields the smallest squared Euclidean distance to $\boldsymbol{f}_s$ during the image generation process, rather than the ``finally-generated'' image produced at the end. Part C in the supplementary materials provides a comparison between these two types of images. The second part presents three example pairs in which the right image is generated based on $\boldsymbol{f}_s$ extracted from the left actual image. Each pair is annotated with the squared Euclidean distance between the generated image and $\boldsymbol{f}_s$, and three human evaluation scores mentioned below.

\begin{figure}[tbp]
	\centering
	\includegraphics[width=\linewidth]{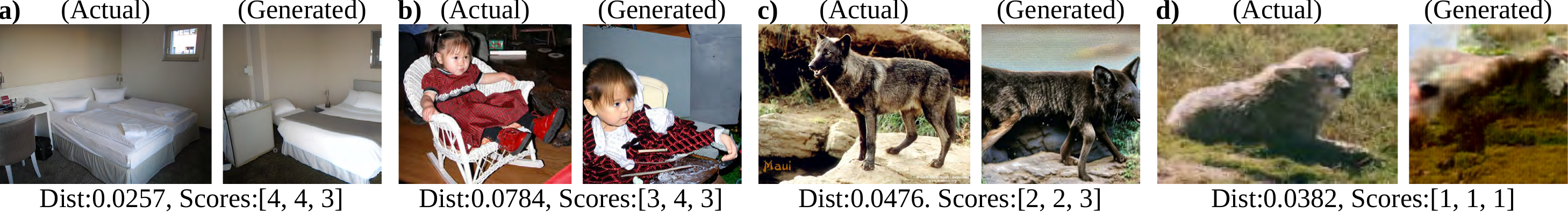}
	\caption{Examples of human evaluation scores. For each generated image, in addition to the squared Euclidean distance between its feature and $\boldsymbol{f}_s$ of the corresponding actual image, the scores given by three human evaluators are shown in square brackets. (a) The generated image acceptably shows a bedroom as in the actual image. (b) The generated image misses a rocking chair that is one of the main objects in the actual image. (c) The anatomical topology of a wolf in the generated image is broken. (d) A grey fox cannot be recognised from the generated image.}
	\label{fig:eval_examples}
	\vspace{-0.5cm}
\end{figure}

 
The last part is a table showing the human evaluation result for the $100$ generated images. Three graduate students in computer science are asked to evaluate these images by selecting one of four scores, as illustrated in Fig.~\ref{fig:eval_examples}:

\noindent \textbf{Acceptable (4):} The main objects, background and overall layout of a generated image are acceptably similar to those of the corresponding actual image. 

\noindent \textbf{Minor unrealism (3):} Although a generated image is acceptably similar to the actual image, the former exhibits minor unrealism, or differs from the latter in one or more of the main objects, background or overall layout of the latter. 
	
\noindent \textbf{Major unrealism but recognisable (2):} A generated image is unrealistic, but its main objects can still be recognised as similar to those of the actual image.

\noindent \textbf{Unrecognisable (1):} A generated image does not allow recognition of any main object in the actual image. 




From the perspective of a feature-matching image generator, the images generated by our decoder are exceptionally similar to the  actual images from which $\boldsymbol{f}_s$ is extracted. To argue this, we search the original (non-sampled) MS-COCO (or ImageNet) dataset to find the nearest actual image to $\boldsymbol{f}_s$. The mean and standard deviation of the squared Euclidean distances between these nearest actual images and $\boldsymbol{f}_s$ are $1.2971 \pm 0.5391$ for the $100$ MS-COCO actual images and $1.2393 \pm 0.5523$ for the $100$ ImageNet actual images. In comparison, the histograms in Fig.~\ref{fig:gen_result_clip} (a) and (b) present that the squared Euclidean distances between the generated images and $\boldsymbol{f}_s$ are characterised by the substantially smaller means and standard deviations. In fact, every generated image has a smaller squared Euclidean distance to $\boldsymbol{f}_s$ than the nearest actual image.


From the perspective of a visual feature space analyser, the human evaluation results in Fig.~\ref{fig:gen_result_clip} (a) and (b) show that the combined proportion of the evaluations ``Acceptable'', ``Minor unrealism'' and ``Major unrealism (recognisable)'' is about $90\%$. This implies that, in most cases, the visual content (at least the main objects) encoded in a feature can be inferred from the feature alone. The mean human evaluation score of approximately $2.5$ indicates that our decoder has room for improvement in generating more realistic images. Meanwhile, this can be considered to reflect imperfections in the feature space of CLIP's image encoder. For example, the rightmost pair in Fig.~\ref{fig:eval_examples} suggests an immature generalisation of the concept ``grey fox'', because the very small squared Euclidean distance between the actual and generated images indicates that a very small change in $\boldsymbol{f}_s$ causes significant unrealism. For the second pair from the right, the broken anatomical topology of the wolf in the generated image may hint that CLIP's image encoder pays little attention to an object's anatomical topology. These observations could be useful for improving CLIP's image encoder.

Using $\boldsymbol{f}_s$ extracted from the same set of $200$ actual images, we run our decoder again to examine its stability against the randomness in its image generation. In this second round, the mean and standard deviation of the squared Euclidean distances between the generated images and $\boldsymbol{f}_s$ are $0.0573 \pm 0.0222$ for the $100$ MS-COCO images and $0.0917 \pm 0.0349$ for the $100$ ImageNet images. Since these values are similar to those in the first round (Fig.~\ref{fig:gen_result_clip} (a) and (b)), our decoder's image generation can be considered relatively stable.

\subsection{Comparative Results Targeting ResNet-50}
\label{ssec:result_resnet}

We compare our training-free decoder to an existing training-based decoder, Representation Conditional Diffusion Model (RCDM)~\cite{F_Bordes}. RCDM is trained by conditioning its image generation on $\boldsymbol{f}_s$ so that a generated image is similar to the actual image from which $\boldsymbol{f}_s$ is extracted. ResNet-50 configured with ResNet50\_Weights.IMAGENET1K\_V1\footnote{\url{https://docs.pytorch.org/vision/main/models/generated/torchvision.models.resnet50.html\#torchvision.models.ResNet50_Weights}} is used as the target feature extractor. For each of the $200$ actual images used in the previous section, $\boldsymbol{f}_s$ is defined as the output of the penultimate layer of this ResNet-50. RCDM employed in this section is actually trained using a large number of images in ImageNet dataset and their features defined in this manner.



\begin{figure}[tbp]
	\centering
	\includegraphics[width=\linewidth]{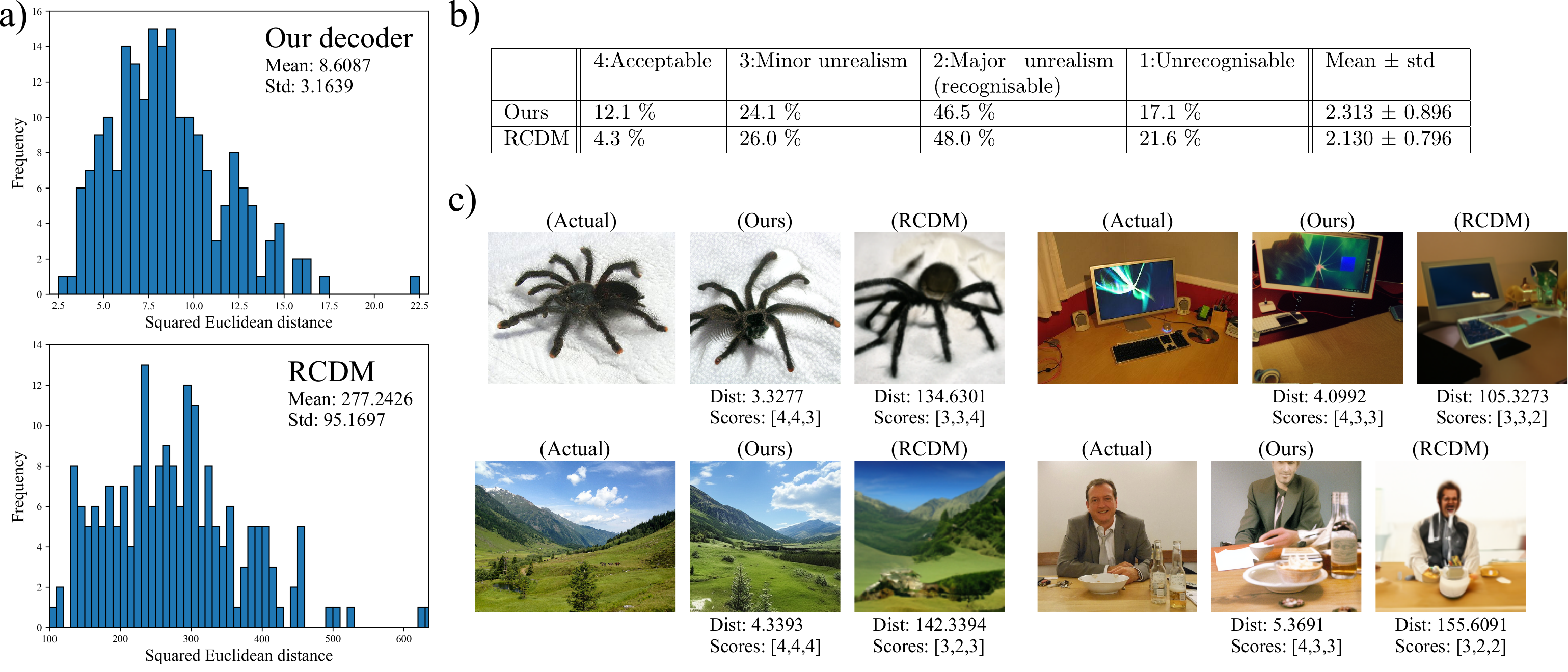}
	\caption{A comparison of images generated by our decoder and RCDM when ResNet-50 is the target feature extractor, and $\boldsymbol{f}_s$ is defined as the feature extracted from each of the $100$ MS-COCO and $100$ ImageNet images used in Fig.~\ref{fig:gen_result_clip}.}
	\label{fig:gen_result_res}
	\vspace{-0.5cm}
\end{figure}

Fig.~\ref{fig:gen_result_res} summarises our experimental comparison between our decoder and RCDM. Due to space limitations, Fig.~\ref{fig:gen_result_res} combines the results for the $100$ MS-COCO and the $100$ ImageNet images. Part D of the supplementary materials provides separate versions of these results for each image set. Fig.~\ref{fig:gen_result_res} (a) presents two histograms, each representing the distribution of the squared Euclidean distances between the $200$ images generated by either our decoder or RCDM and the corresponding $\boldsymbol{f}_s$. Fig.~\ref{fig:gen_result_res} (b) compares our decoder to RCDM in terms of the human evaluation scores assigned to their generated images. As a qualitative comparison, Fig.~\ref{fig:gen_result_res} (c) shows four triples in which the centre and right images are generated by our decoder and RCDM, respectively, when $\boldsymbol{f}_s$ is defined as the feature extracted from the left actual image. Each generated image is annotated with the squared Euclidean distance between its feature and $\boldsymbol{f}_s$, and its three human evaluation scores.


\begin{figure}[tbp]
	\centering
	\includegraphics[width=\linewidth]{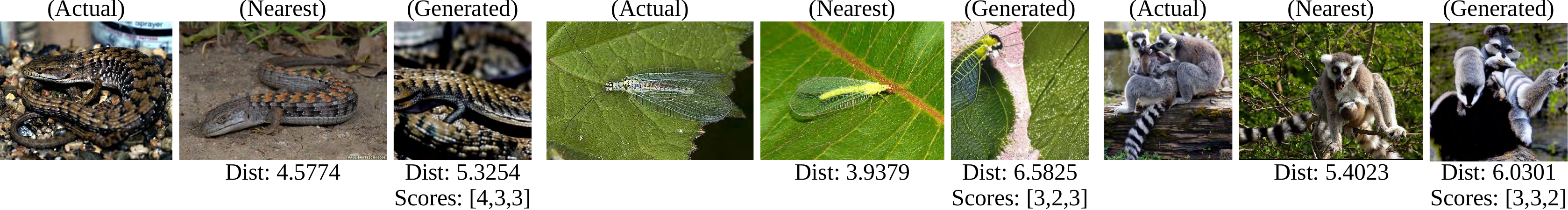}
	\caption{Three of the five cases where an image generated by our decoder has a larger squared Euclidean distance to $\boldsymbol{f}_s$ (extracted from an actual ImageNet image) than the nearest image retrieved from the non-sampled ImageNet dataset.}
	\label{fig:lost_cases}
	\vspace{-0.5cm}
\end{figure}

Fig.~\ref{fig:gen_result_res} demonstrates that our decoder generates images whose features are considerably closer to $\boldsymbol{f}_s$ than RCDM. Fig.~\ref{fig:gen_result_res} (a) shows that the mean and standard deviation of the squared Euclidean distances between the generated images and $\boldsymbol{f}_s$ are $8.6087 \pm 3.1639$ and $277.246 \pm 95.1697$ for our decoder and RCDM, respectively. Similar to the previous section, we interpret these values by computing the mean and standard deviation of the squared Euclidean distances between $\boldsymbol{f}_s$ for the $200$ actual images and their nearest images retrieved from the non-sampled MS-COCO (or ImageNet) dataset. The resulting value, $70.6619 \pm 53.4817$, confirms the closeness of our decoder's generated images to $\boldsymbol{f}_s$. Moreover, as exemplified in Fig.~\ref{fig:lost_cases}, there are five cases where a generated image fails to achieve a smaller squared Euclidean distance to $\boldsymbol{f}_s$ than the nearest image. In each of these five cases, the actual image and its nearest image belong to the same ImageNet class. One plausible explanation is that ResNet-50 is trained on ImageNet dataset to classify these images into the same class, so their features tend to be highly similar. In other words, only a small fraction of such exceptional images lie closer to $\boldsymbol{f}_s$ than our decoder's generated images.


Fig.~\ref{fig:gen_result_res} (b) suggests that our decoder serves as a better visual feature space analyser than RCDM in terms of human evaluation scores. In particular, compared to RCDM, images generated by our decoder tend to maintain details of the corresponding actual images such as hairs and red parts of a tarantula's legs, light displayed on a computer monitor, trees in the foreground, and a dish and bottles on a table, as illustrated in Fig.~\ref{fig:gen_result_res} (c). Finally, the proportion of the human evaluation ``Unrecognisable'' in Fig.~\ref{fig:gen_result_res} (b) ($17.1\%$) is higher than those in Fig.~\ref{fig:gen_result_clip} (a) and (b) ($6.3\%$ and $13.6\%$). Apart from the room for improvement in our decoder, this may indicate that ResNet-50's feature space has more imperfections than that of CLIP's image encoder. That is, the quality of generated images may be linked to the performance of the target feature extractor. 

\subsection{Discussions}
\label{ssec:discussions}


\noindent \textbf{Improving the realism of generated images:} Considering that the largest proportion of human evaluation scores is assigned to ``Major unrealism but recognisable'', our decoder has much room for improvement in generating more realistic images. We hypothesise that unrealistic content may waste image regions that could otherwise be used to depict content relevant to a user-specified feature. Thus, generating a more realistic image may make its feature closer to the user-specified one. To this end, we plan to enhance the loss function in Eq. (\ref{eq:euc_loss}) by incorporating an image quality assessment~\cite{R_Zhang,J_Wang} of the generated image at each step. In addition, we will formulate our decoder's image generation as a search problem to identify better noises to be injected~\cite{N_Ma}. Also, we will investigate the possibility of detecting unrealism by examining whether the latent representation at each step lies within the support of the distribution of a pre-trained diffusion model~\cite{Z_Tang}, and by monitoring the convergence of features extracted from images generated at successive steps using a held-out feature extractor.



\noindent \textbf{Recuding our decoder's computation time:} One drawback of our decoder is its computation time. The average computation times over $45$ image generation runs on NVIDIA GeForce RTX 3090 are $8830$ and $8951$ seconds when targeting CLIP's image encoder with ResNet-50 backbone and ResNet-50, respectively. One approach to reduce these times is to adopt a fast sampler that reduces the number of steps required for image generation of a pre-trained diffusion model. Our preliminary experiment shows that popular DDIM sampler~\cite{J_Song} does not work well, which may be attributed to poor discretisation of the continuous time underlying the image generation process into discrete steps. Thus, we will explore an approach to learn the optimal time discretisation~\cite{V_Tong}. By the way, our decoder consumes about $15.1$ and $14.9$ GB of VRAM for image generation targeting CLIP's image encoder and ResNet-50, respectively.


\noindent \textbf{Smooth alignment between features and visual appearances:} As demonstrated before, our decoder successfully generates an image that is located in close proximity to an actual image in the feature space of a target feature extractor. Although the generated image captures the main content of the actual image, it also exhibits unrealistic and unnatural visual appearances. In other words, even a small deviation from the actual image in the feature space can result in substantial degradation of visual realism, as illustrated by the rightmost generated images in Figs.~\ref{fig:eval_examples} and \ref{fig:lost_cases}. Therefore, a potential direction for improving current feature extractors is to design a training objective that enforces smooth transitions in visual appearances of images according to their Euclidean distances in the feature space.


\noindent \textbf{Instance-level visualisation of CLIP's modality gap:} CLIP~\cite{A_Radford} is a visual-semantic embedding model that projects images and captions into a common feature (embedding) space. The \textit{modality gap} refers to the phenomenon that images and captions are located in completely different subspaces within such a multimodal feature space\cite{W_Liang,S_Schrodi}. Below, unlike existing studies that mainly focus on theoretical analysis and feature-distribution visualisation~\cite{W_Liang,S_Schrodi}, our decoder is utilised to perform an ``instance-level'' visualisation, revealing the absence of images around captions in CLIP's feature space. 


\begin{figure}[htb]
	\centering
	\includegraphics[width=\linewidth]{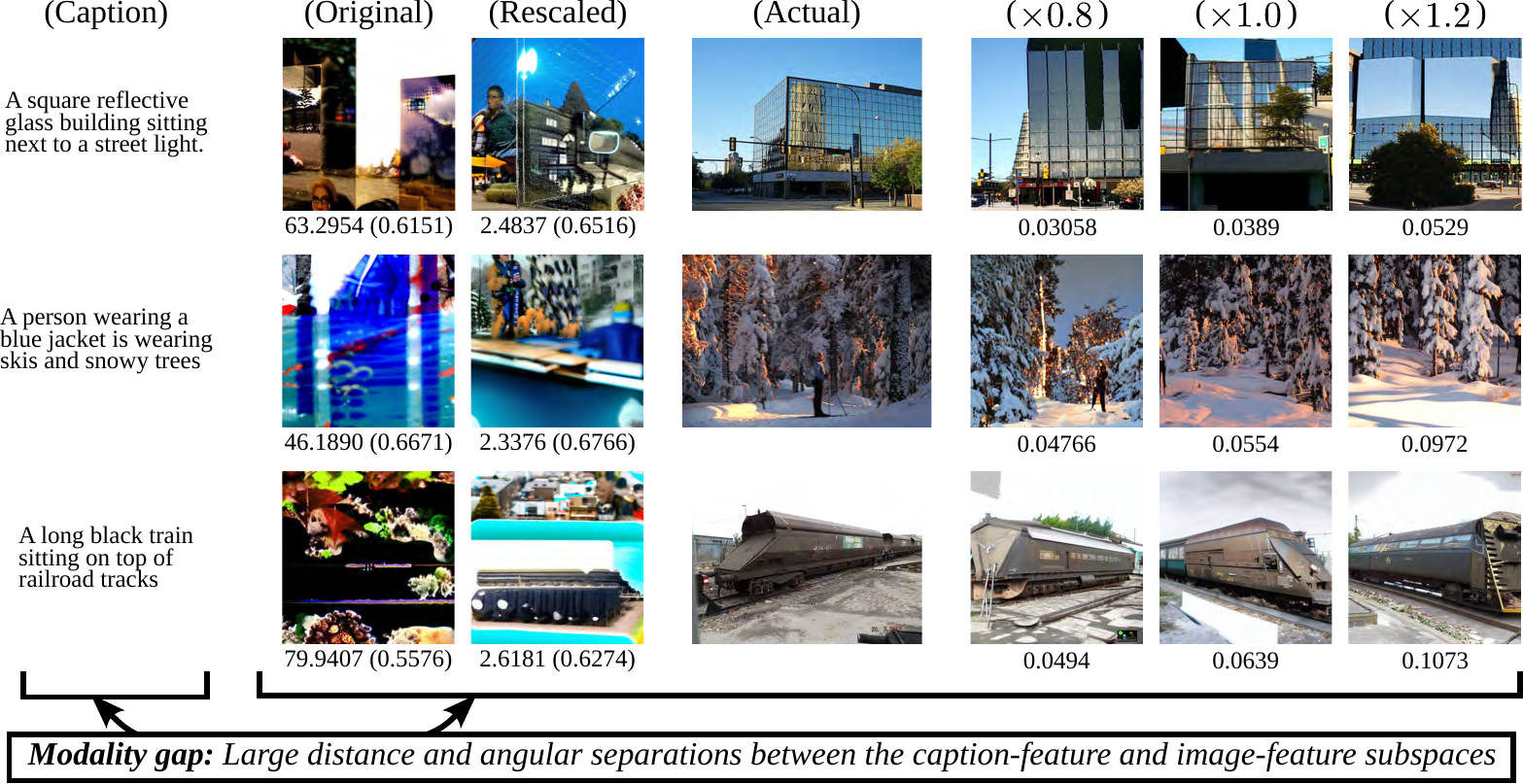}
	\caption{Instance-level visualisation of the modality gap in CLIP's feature space. Each row is based on a paired caption and image shown in the first and fourth columns, respectively. The second and third columns present images generated by defining $\boldsymbol{f}_s$ as a caption feature extracted by CLIP's text encoder and its rescaled version, respectively. The rightmost three columns display images generated by defining $\boldsymbol{f}_s$ as $0.8\times$-, $1.0\times$- (i.e., original) and $1.2\times$-scaled versions of an actual image feature extracted by CLIP's image encoder. The number under each generated image and the number in brackets indicate the squared Euclidean distance and normalised cosine similarity between its feature and $\boldsymbol{f}_s$ used to generate it, respectively. When $\boldsymbol{f}_s$ is an image feature, the cosine similarity between $\boldsymbol{f}_s$ and the feature of every generated image is always larger than $0.99$, so such cosine similarities are omitted.}
	\label{fig:gen_result_t2i}
	\vspace{-0.5cm}
\end{figure}

We pair each of the $100$ MS-COCO images used before with the first of the five captions assigned to it. The ``Caption'' and ``Actual'' columns in Fig.~\ref{fig:gen_result_t2i} show three image-caption pairs. We examine whether an image generated by defining $\boldsymbol{f}_s$ as the feature of a caption is similar to the corresponding actual image. As shown in the ``Original'' column, the generated images are completely different from the actual images, and their features have very large squared Euclidean distances to the features of the corresponding captions. In particular, the mean and standard deviation of these distances for the $100$ image-caption pairs are $66.2885 \pm 18.8549$ (min: $32.7989$, max: $134.3457$).

CLIP is trained to encode an image and its paired caption into similar features with respect to their (normalised) cosine similarities~\cite{A_Radford}, not their squared Euclidean distances. In this regard, the mean and standard deviation of the cosine similarities for the $100$ image-caption pairs are $0.2509 \pm 0.0357$ (max: $0.3289$, min: $0.1262$)\footnote{The mean and standard deviation of the cosine similarities between non-paired images and captions are $0.1005 \pm 0.0411$ (max: $0.2951$, min: $-0.0731$)}. In comparison, the cosine similarities between the features of our decoder's generated images and those of the corresponding captions are significantly higher, namely $0.6028 \pm 0.0675$ (min: $0.3435$, max: $0.7613$). This means that, in CLIP's feature space, the generated images are oriented much closer to the corresponding captions than the paired actual images. Nonetheless, they are located in substantially different subspaces of CLIP's feature space, as supported by the fact that the average feature norms of the $100$ actual images and the paired captions are $2.0770$ and $10.1094$, respectively. 

We further run our decoder by defining $\boldsymbol{f}_s$ as a rescaled caption feature so that its norm is equal to the average feature norm of the $100$ actual images. However, the ``Rescaled'' column in Fig.~\ref{fig:gen_result_t2i} shows that the generated images still differ significantly from the actual images, although they slightly align with the captions, as objects described therein (e.g., building, person, and black train) become barely recognisable. Compared to the case without rescaling, the mean and standard deviation of the squared Euclidean distances between the generated image features and the rescaled caption features are substantially reduced to $2.4795 \pm 0.3119$ (min: $1.7471$, max: $3.1865$), whereas only a modest improvement is observed on the mean and standard deviation of their cosine similarities, namely $0.6502 \pm 0.0560$ (min: $0.5113$, max: $0.7713$).


The above failure to generate reasonable images is unlikely to be attributable to our decoder. The columns named ``$1.2 \times$'' and ``$0.8 \times$'' present images generated by setting $\boldsymbol{f}_s$ to $1.2$ and $0.8$ times the feature of an actual image, respectively. Since both the original and scaled features have the same normalised cosine similarity to the paired caption feature, we hypothesise that an image generated by our decoder to closely match a scaled feature should be semantically relevant to the caption. The generated images in Fig.~\ref{fig:gen_result_t2i} appear to support this hypothesis. Thus, as long as there exist real images whose features lie near a user-specified feature in the feature space, our decoder is able to generate reasonable images that capture their main contents. Therefore, we conclude that the large Euclidean distances between the caption features--even after rescaling--and the generated image features suggest the absence of real images near those caption features (i.e., modality gap). Although these distances could be reduced to some extent by adopting a more advanced pre-trained diffusion model than stable diffusion, our conclusion is likely to remain valid.


Our instance-level visualisation provides a possible direction for addressing the modality gap. Assuming that the speed of our decoder could be sufficiently improved, the modality gap could be reduced by devising a loss function that examines whether our decoder can generate images that not only have features close to the paired caption features but are also realistic and caption-relevant.


\section{Conclusion and Future Work}
\label{sec:conclusion}


To enable precise analysis of a target feature extractor's feature space, this paper introduces a decoder that generates images whose features closely match user-specified ones. Our decoder is implemented as a training-free guided diffusion model that guides a pre-trained diffusion model (stable diffusion) to minimise the squared Euclidean distance between a user-specified feature and that of a generated image. The experiments using CLIP's image encoder with ResNet-50 backbone and ResNet-50 as the target feature extractors demonstrate that our decoder can not only generate images whose features are remarkably similar to user-specified ones, but also function as an effective visual feature space analyser, as the primary visual contents encoded in the specified features can be recognised from most generated images.

In addition to the future directions discussed in Section~\ref{ssec:discussions}, we plan to implement our decoder with pre-trained diffusion models (e.g., SVG~\cite{M_Shi} and PixelDiT~\cite{Y_Yu}) other than stable diffusion, analyse the feature spaces of more recent feature extractors than those employed in this paper (e.g., image encoders implemented in OpenCLIP project~\cite{G_Ilharco}), and extend our decoder to investigate the feature spaces of other modalities such as text, audio and time-series. 


%
%
\bibliographystyle{splncs04}
\bibliography{refs}


\newpage

\makeatletter
\renewcommand{\@seccntformat}[1]{%
	\ifcsname the#1\endcsname
	Part~\csname the#1\endcsname\quad
	\fi}
\makeatother

\renewcommand{\thesection}{\Alph{section}}
\renewcommand{\thesubsection}{\thesection.\alph{subsection}}

\setcounter{section}{0}
\setcounter{figure}{0}
\renewcommand{\figurename}{Fig.}
\renewcommand{\thefigure}{S\arabic{figure}}

\renewcommand{\thetable}{S\arabic{table}}

\section*{Supplementary Materials}

Please note that the figures in the main paper are numbered like ``Fig.~$1$'', ``Fig.~$2$'' and so on, while the figures in the supplementary materials are numbered with an ``S'' prefix like ``Fig.~S$1$'', ``Fig.~S$2$'' and so on. In addition, to distinguish the sections in the main paper from those in the supplementary materials, the latter are labeled alphabetically as ``Part A'', ``Part B'' and so on.


\section{Distribution of Values in $\nabla_{\boldsymbol{z}_t} l \left( f \left( \hat{\boldsymbol{x}}_t \right), \boldsymbol{f}_s \right)$}

\begin{sidewaysfigure}
	\centering
	\includegraphics[width=\linewidth]{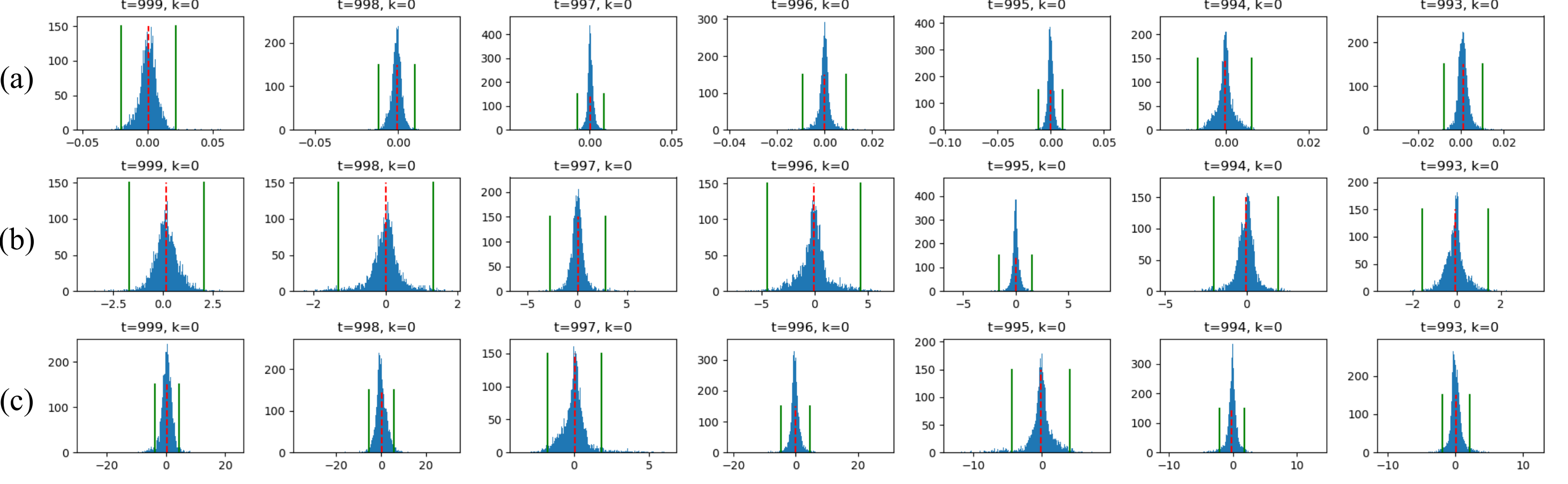}
	\caption{Distributions of values in $\nabla_{\boldsymbol{z}_t} l \left( f \left( \hat{\boldsymbol{x}}_t \right), \boldsymbol{f}_s \right)$ at the first self-recurrence iteration (i.e., $k=0$) for each of the first seven reverse steps (i.e., $t=999,\cdots, 993$) when the target feature extractor $f$ is defined as CLIP's image encoder with ResNet-50 backbone (a), ResNet-50 (b) and ViT-H/14 (c). Here, $\boldsymbol{f}_s$ is defined as the feature extracted from image $1$ in Fig.~\ref{fig:act_imgs}. In each visualised distribution, the dotted line indicates the mean and the solid lines represent $\pm \nabla_{thres}$ used in gradient clipping (i.e., three times the standard deviation for (a) and (b), and twice the standard deviation for (c)).}
	\label{fig:grad_dist}
\end{sidewaysfigure}

The gradient clipping described in Section~$3$ is motivated by our observation that values in $\nabla_{\boldsymbol{z}_t} l \left( f \left( \hat{\boldsymbol{x}}_t \right), \boldsymbol{f}_s \right)$ approximately follow a Gaussian distribution whose mean is nearly zero as demonstrated in Fig.~\ref{fig:grad_dist}. Notably, the plots in Fig.~\ref{fig:grad_dist} (c) where ViT-H/14 (described in the next part) is the target feature extractor span much larger horizontal ranges than those in Fig.~\ref{fig:grad_dist} (a) and (b) where CLIP's image encoder with ResNet-50 backbone and ResNet-50 are targeted respectively. That is, for Fig.~\ref{fig:grad_dist} (c), values in $\nabla_{\boldsymbol{z}_t} l \left( f \left( \hat{\boldsymbol{x}}_t \right), \boldsymbol{f}_s \right)$ are distributed over much wider ranges. Thus, when targeting ViT-H/14, $\nabla_{thres}$ is set to twice the standard deviation of values in $\nabla_{\boldsymbol{z}_t} l \left( f \left( \hat{\boldsymbol{x}}_t \right), \boldsymbol{f}_s \right)$ in order to filter out a greater proportion of overly large values. In comparison, a larger multiplier of $3$ is used when targeting CLIP's image encoder with ResNet-50 backbone and ResNet-50 so as to retain a larger proportion of values in $\nabla_{\boldsymbol{z}_t} l \left( f \left( \hat{\boldsymbol{x}}_t \right), \boldsymbol{f}_s \right)$.


\section{Additional Results Targeting ViT-14/L}

\begin{figure}[htbp]
	\centering
	\includegraphics[width=\linewidth]{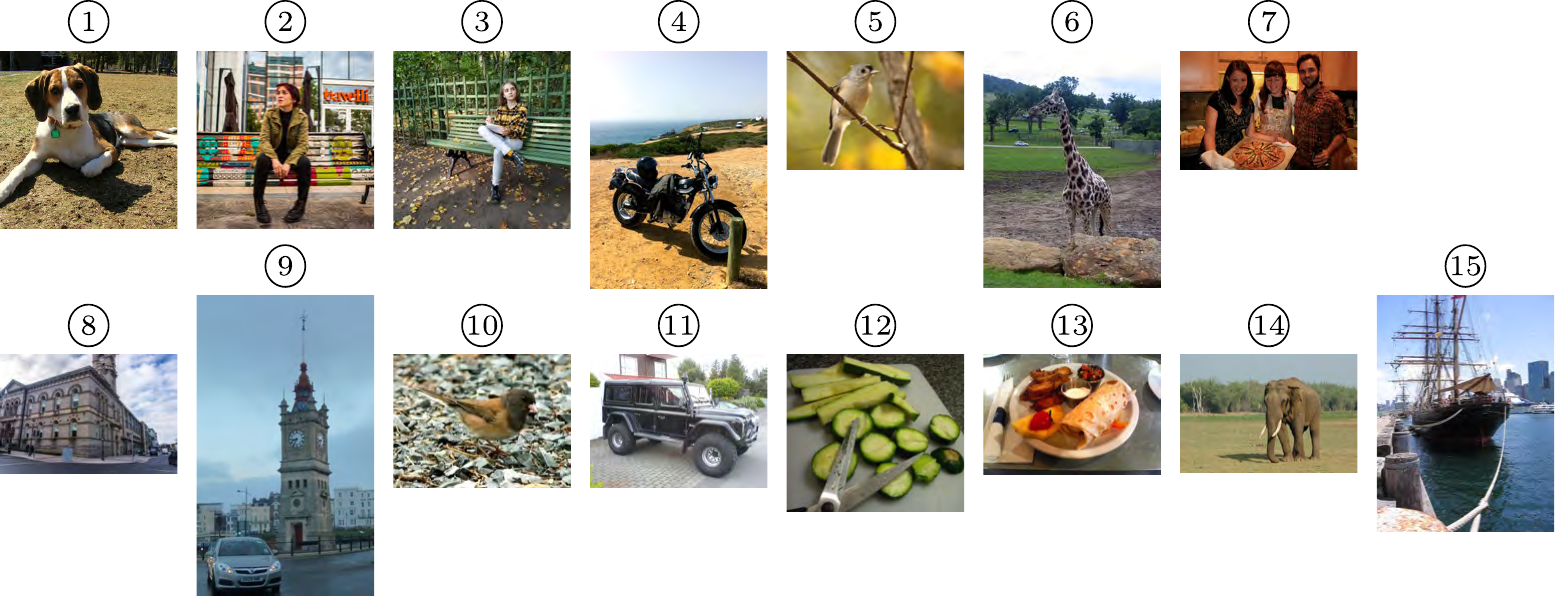}
	\caption{$15$ actual images used in the additional experiments.}
	\label{fig:act_imgs}
\end{figure}

To check the generality of our decoder, we conduct an additional experiment using Vision Transformer (ViT)~\cite{A_Dosovitskiy} as the target feature extractor. In particular, ViT-H/14 configured with ViT\_H\_14\_Weights. IMAGENET1K\_SWAG\_E2E \_V1~\footnote{\url{https://docs.pytorch.org/vision/main/models/generated/torchvision.models.vit_h_14.html\#torchvision.models.ViT_H_14_Weights}} is selected because it is ranked at the top in the ranking of the pre-trained models in Pytorch~\footnote{\url{https://docs.pytorch.org/vision/stable/models.html\#table-of-all-available-classification-weights} (accessed on July 5, 2026)}. In this small-scale additional experiment, $\boldsymbol{f}_s$ is extracted from each one of the $15$ actual images shown in Fig.~\ref{fig:act_imgs}. Images $1$-$3$ are drawn from the dataset used in \cite{A_Bansal,A_Bansal2}, images $4$-$9$ are from MS-COCO dataset~\cite{T_Lin}, and the remaining images are from ImageNet dataset~\cite{O_Russakovsky}. All the hyper-parameters are the same to the experiments in the main paper, except for $\nabla_{thres}$ that is set to twice the standard deviation of values in $\nabla_{\boldsymbol{z}_t} l \left( f \left( \hat{\boldsymbol{x}}_t \right), \boldsymbol{f}_s \right)$ due to the broad spread of these values, as illustrated in Fig.~\ref{fig:grad_dist}.

\begin{figure*}[htbp]
	\centering
	\includegraphics[width=\linewidth]{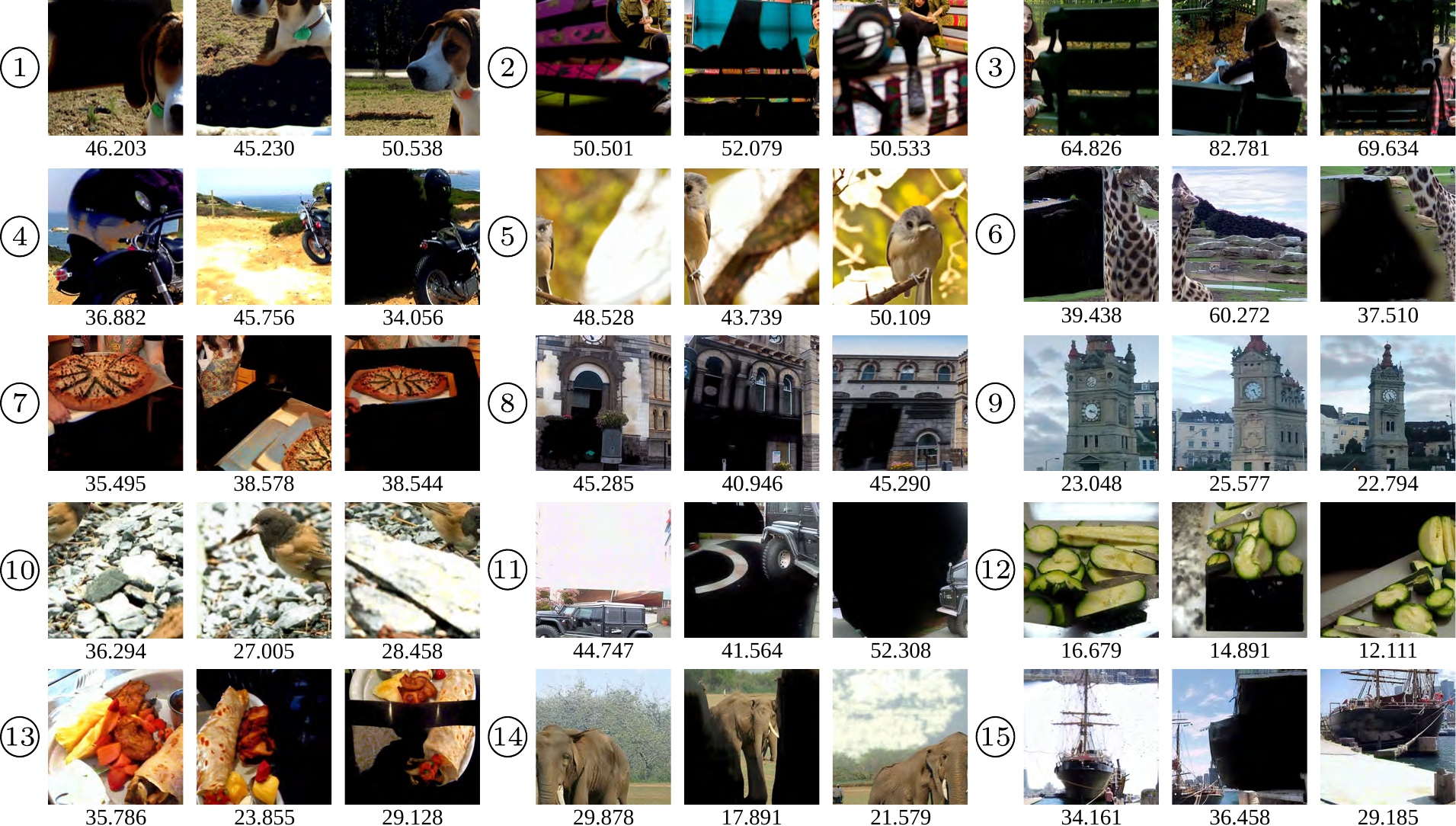}
	\caption{Images generated by our decoder when using ViT-H/14 as the target feature extractor. For each index, three generated images are shown together with the squared Euclidean distances between their features and that of correspondingly-indexed actual image in Fig.~\ref{fig:act_imgs}.}
	\label{fig:gen_result_vit}
\end{figure*}

Fig.~\ref{fig:gen_result_vit} shows images generated by our decoder when using ViT-14/H as the target feature extractor. For each index in Fig.~\ref{fig:gen_result_vit}, $\boldsymbol{f}_s$ is defined as the feature of the correspondingly-indexed actual image in Fig.~\ref{fig:act_imgs}, and our decoder is run three times by considering the randomness in its image generation. Like the experiments in the main paper, each run is represented by the best image that attains the smallest squared Euclidean distance to $\boldsymbol{f}_s$ during the image generation process. 

\begin{figure}[htbp]
	\centering
	\includegraphics[width=\linewidth]{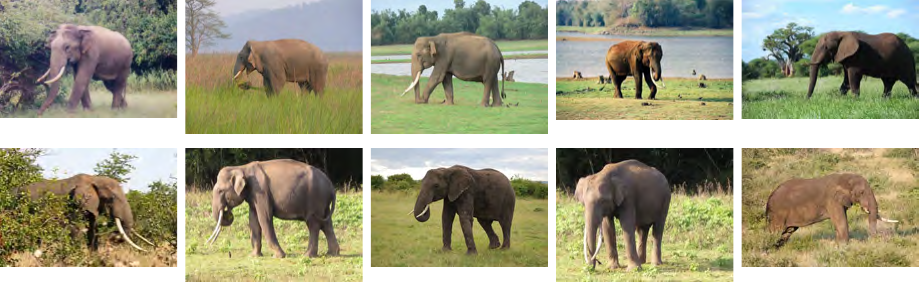}
	\caption{$10$ images that are carefully chosen from the tusker category of ImageNet dataset~\cite{O_Russakovsky} and validated by human judgment to be visually similar.}
	\label{fig:tusker_images}
\end{figure}

As a reference to interpret the generated images in Fig.~\ref{fig:gen_result_vit}, the average pairwise squared Euclidean distance among the $15$ actual images in Fig.~\ref{fig:act_imgs} is $2914.379$. In addition, the $10$ images in Fig.~\ref{fig:tusker_images} are carefully chosen from the tusker category of ImageNet dataset and confirmed to have similar visual perception by human. The average pairwise squared Euclidean distance among these $10$ tusker images is $715.504$. By taking these values as a reference, it is suggested that our decoder successfully generates images whose features are remarkably similar to $\boldsymbol{f}_s$.

The images generated by targeting ViT-H/14 in Fig.~\ref{fig:gen_result_vit} seem to preserve detailed contents of the corresponding actual images. For example, for index $1$, the dogs in the generated images in Fig.~\ref{fig:gen_result_vit} have objects around their necks (although one of them is red). For index $7$ in Fig.~\ref{fig:gen_result_vit}, the pizzas in the generated images closely resemble that in the actual image in Fig.~\ref{fig:act_imgs}. Moreover, as seen for index $15$ in Fig.~\ref{fig:gen_result_vit}, both the generated images and the actual image contain tall buildings in the background (although they appear to be drawn in the sky in the left generated image).

One typical artifact observed in the generated images in Fig.~\ref{fig:gen_result_vit} is the presence of large black regions. When a large portion of an image is occupied by such a black region, only the remaining region can be used to make its feature match a user-specified feature, which significantly limits the possibility of images that can be generated. The approaches discussed in Section~$4.3$ of the main paper to improve the realism of our decoder's generated images are expected to be useful for mitigating artifactual black regions. 

\section{Comparison between the Best and Finally-generated Images}
\label{sec:best_vs_final}

Figs.~\ref{fig:best_vs_final_1} and \ref{fig:best_vs_final_2} show a comparison between ``finally-generated'' and ``best'' images when using CLIP's image encoder with ResNet-50 backbone as the target feature extractor. For each run, the finally-generated image is obtained at the end, whereas the best image is the one whose feature yields the smallest squared Euclidean distance to the user-specified feature $\boldsymbol{f}_s$ during the image generation process. In Figs.~\ref{fig:best_vs_final_1} and \ref{fig:best_vs_final_2}, $\boldsymbol{f}_s$ is defined as the feature of the correspondingly-indexed actual image in Fig.~\ref{fig:act_imgs}, and our decoder is run three times. The number under each generated image shows the squared Euclidean distance between its feature and $\boldsymbol{f}_s$. The comparison between the finally-generated and best images indicates that both are visually very similar. However, enriching the best image with additional details to obtain the finally-generated image degrades its squared Euclidean distance and visual similarity to the actual image.

\begin{figure}[htbp]
	\centering
	\includegraphics[width=\linewidth]{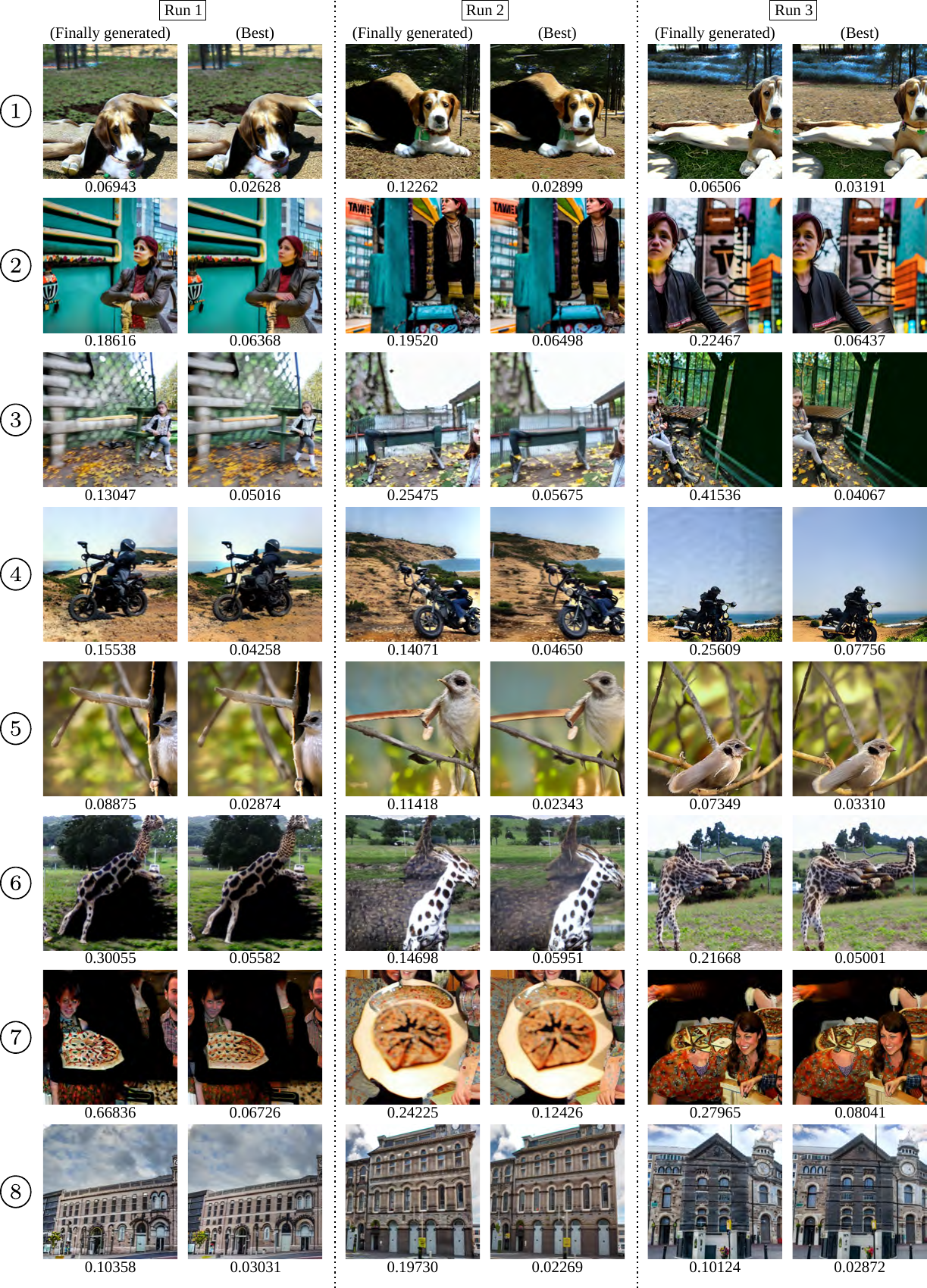}
	\caption{A comparison between finally-generated and best images when using CLIP's image encoder with ResNet-50 backbone as the target feature extractor. For each index, our decoder is run three times to generate images whose features closely match the feature of the correspondingly-indexed image in Fig.~\ref{fig:act_imgs}. For each run, the finally-generated image and the best image are shown together with the squared Euclidean distances between their features and that of the corresponding actual image.}
	\label{fig:best_vs_final_1}
\end{figure}

\begin{figure}[htbp]
	\centering
	\includegraphics[width=\linewidth]{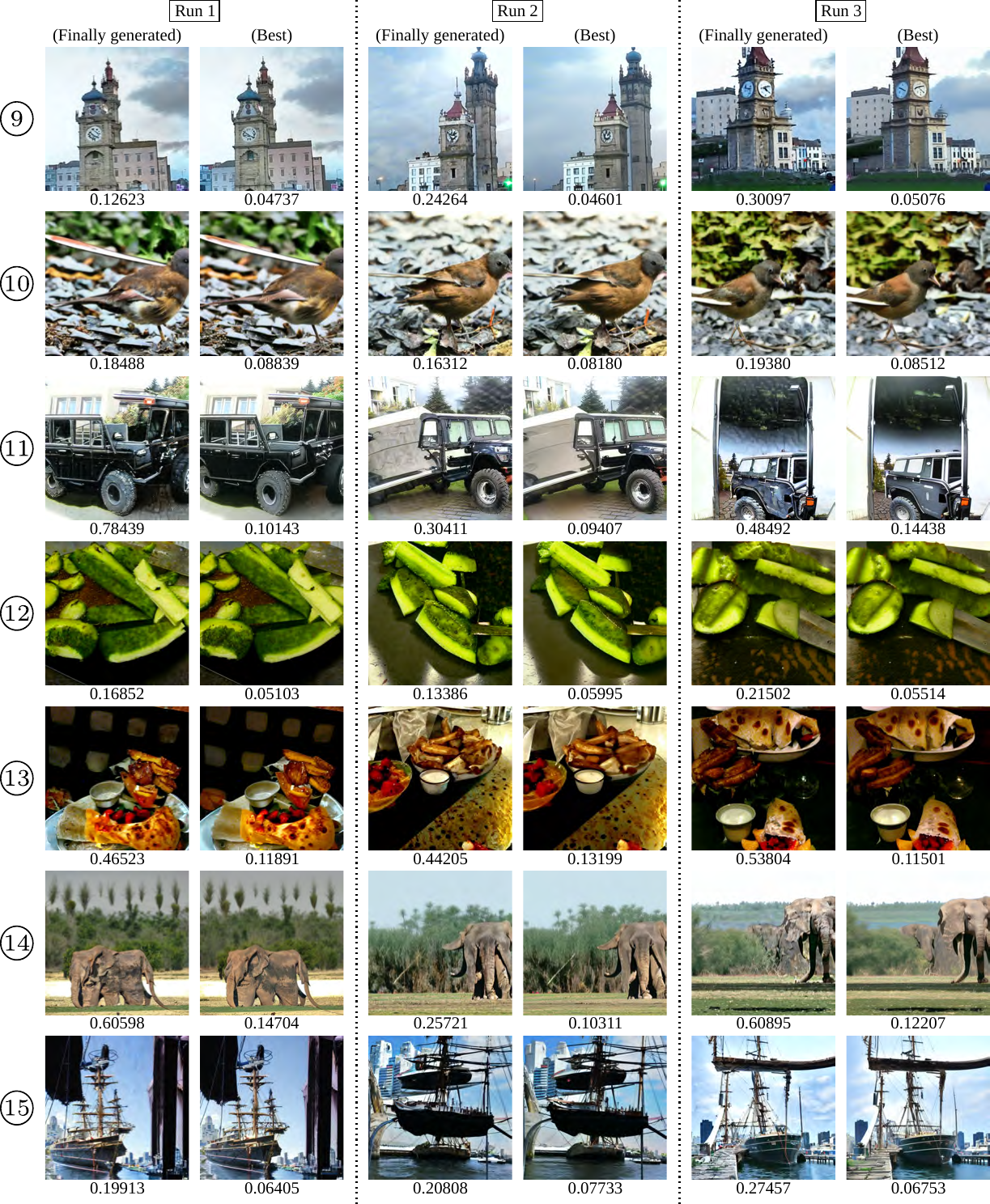}
	\caption{(Continued from Fig.~\ref{fig:best_vs_final_1}) A comparison between finally-generated and best images when using CLIP's image encoder with ResNet-50 backbone as the target feature extractor.}
	\label{fig:best_vs_final_2}
\end{figure}

Also, in Section $4.1$ of the main paper, $\boldsymbol{f}_s$ is defined as the feature extracted from each of the $100$ MS-COCO images and $100$ ImageNet images using CLIP's image encoder with ResNet-50 backbone. In this setting, the mean and standard deviation of the squared Euclidean distances between the features of the finally-generated images and $\boldsymbol{f}_s$ are $0.2533 \pm 0.1354$, whereas those for the best images are $0.0764 \pm 0.0331$. In Section $4.2$ where $\boldsymbol{f}_s$ is defined by applying ResNet-50 to the same set of $200$ images, these statistics for the finally-generated and best images are $24.4389 \pm 12.3169$ and $8.6087 \pm 3.1639$, respectively. 

We conjecture that the above-mentioned non-smooth convergence in our decoder's image generation results from using a constant guidance weight $w_g$ across all the steps. We will investigate an approach to dynamically tune $w_g$ based on the status of the image generated (more precisely, the predicted clean image) at each step.


\section{Detailed Results Targeting ResNet-50}

\begin{figure}[htbp]
	\centering
	\includegraphics[width=\linewidth]{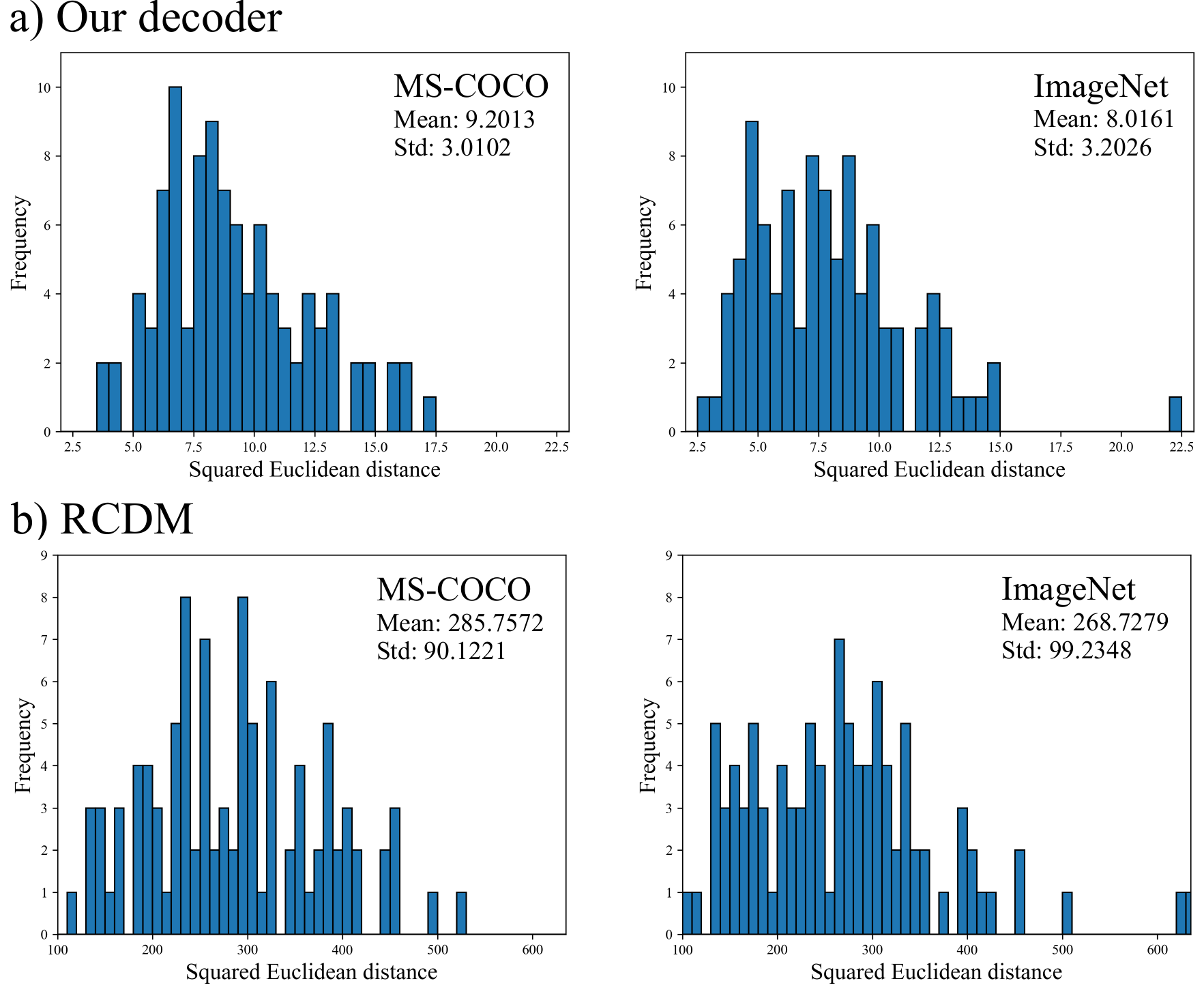}
	\caption{Comparison of histograms of the squared Euclidean distances between the user-specified features derived from the $100$ MS-COCO (or $100$ ImageNet) images and the features of the images generated by our decoder (a) and RCDM (b), with ResNet-50 used as the target feature extractor.}
	\label{fig:resnet_separate}
\end{figure}

Due to space limitations, Section $4.2$ of the main paper reports combined results for the $100$ MS-COCO and $100$ ImageNet images. For completeness, Fig.~\ref{fig:resnet_separate} shows, for both our decoder and RCDM, two histograms each of which represents the distribution of the squared Euclidean distances between the features of the generated images and the corresponding MS-COCO or ImageNet image features (i.e., user-specified features $\boldsymbol{f}_s$). In addition, Table~\ref{tbl:comp_res} presents the human evaluation results for the $100$ images generated by our decoder and RCDM, separately when $\boldsymbol{f}_s$ is defined using individual MS-COCO and ImageNet images.

Similar to Fig. $4$ (a) in the main paper, Fig.~\ref{fig:resnet_separate} confirms that our decoder generates images whose features are considerably closer to user-specified features than RCDM. Table~\ref{tbl:comp_res} shows that our decoder is significantly superior to RCDM when user-specified features are extracted from the $100$ MS-COCO images, whereas these methods are comparable for the $100$ ImageNet images. One possible reason is that the ImageNet case is favourable for RCDM because it was trained on ImageNet images~\cite{F_Bordes}\footnote{\url{https://github.com/facebookresearch/RCDM/blob/main/model-card.md} (accessed on October 3, 2026)}. Nevertheless, our decoder achieves comparable human evaluation scores. 

{
	\setlength{\tabcolsep}{10pt}
	
	\begin{table}[htbp]
		\caption{Comparison of human evaluation scores for images generated by specifying features extracted from the $100$ MSCOCO images (a) and $100$ ImageNet images (b) through ResNet-50.}
		\label{tbl:comp_res}
		\centering
		\begin{tabular}{|c||c|c|c|c||c|} 
			\multicolumn{6}{c}{\textbf{(a) MS-COCO}} \\ \hline
			Score & 4 & 3 & 2 & 1 & Mean $\pm$ std \\ \hline
			Ours & 8.6 \%& 24.0 \% & 48.3 \% & 19.0 \% & 2.223 $\pm$ 0.853 \\ \hline
			RCDM & 0.3 \%& 16.0 \% & 53.0 \% & 30.6 \% & 1.860 $\pm$ 0.679 \\ \hline
			\multicolumn{6}{c}{ }\\
			\multicolumn{6}{c}{\textbf{(b) ImageNet}} \\ \hline
			Score & 4 & 3 & 2 & 1 & Mean $\pm$ std \\ \hline
			Ours & 15.6 \%& 24.3 \% & 44.6 \% & 15.3 \% & 2.403 $\pm$ 0.929 \\ \hline
			RCDM & 8.3 \%& 36.0 \% & 43.0 \% & 12.6 \% & 2.400 $\pm$ 0.813 \\ \hline
		\end{tabular}
	\end{table}
	
}

\end{document}